\documentclass[letterpaper]{article} 
\pdfoutput=1
\usepackage{aaai24}  
\usepackage{times}  
\usepackage{helvet}  
\usepackage{courier}  
\usepackage[hyphens]{url}  
\usepackage{graphicx} 
\usepackage{natbib}  
\usepackage{caption} 
\usepackage{algorithm}
\usepackage{algorithmic}

\usepackage{newfloat}
\usepackage{listings}
\DeclareCaptionStyle{ruled}{labelfont=normalfont,labelsep=colon,strut=off} 
\floatstyle{ruled}
\newfloat{listing}{tb}{lst}{}
\floatname{listing}{Listing}
\usepackage{subfigure}
\usepackage{multirow}
\usepackage{lipsum}
\usepackage{rotating}

\usepackage{multirow}
\usepackage[table,xcdraw]{xcolor}

\usepackage[table]{xcolor}

\definecolor{darkgrn}{rgb}{0, 0.75, 0}

\definecolor{aliceblue}{rgb}{0.94, 0.97, 1.0}
\definecolor{bittersweet}{rgb}{1.0, 0.44, 0.37}
\definecolor{mgreen}{rgb}{0.0, .81, .51}
\colorlet{shadecolor}{gray!40}
\RequirePackage{booktabs}
\usepackage[capitalise]{cleveref}

\usepackage{xargs}   
\usepackage{xspace}

\usepackage[textsize=scriptsize,textwidth=25mm]{todonotes}

\newcommandx{\ak}[2][1=]{\todo[color=green!50,#1]{\sf \textbf{AK:} #2}\xspace}

\newcommandx{\kt}[2][1=]{\todo[color=blue!50,#1]{\sf \textbf{KT:} #2}\xspace}

\newcommandx{\kr}[2][1=]{\todo[color=green!50,#1]{\sf \textbf{Kirti:} #2}\xspace}

\newcommandx{\af}[2][1=]{\todo[color=green!50,#1]{\sf \textbf{Alan:} #2}\xspace}

\newcommandx{\bs}[2][1=]{\todo[color=green!50,#1]{\sf \textbf{Bhupi:} #2}\xspace}

\newcommandx{\supriya}[2][1=]{\todo[color=green!50,#1]{\sf \textbf{Supriya:} #2}\xspace}

\newcommand{\snotel}{SNOTEL}

\newcommand{\inputseqx}{\textbf{x}$^{j,h}$}
\newcommand{\outputseqy}{\textbf{y}$^{j,h}$}
\newcommand{\embedseqe}{\textbf{e}$^{j,h}$}
\newcommand{\encodseqa}{\textbf{a}$^{j,h}$}
\newcommand{\concatseqae}{\textbf{ae}$^{j,h}$}
\newcommand{\reduceddimz}{\textbf{z}$^{j,h}$}

\newcommand{\spaceatt}{\texttt{Spatial\_att}}
\newcommand{\temporalatt}{\texttt{Temporal\_att}}
\newcommand{\ensembleatt}{\texttt{Ensemble\_att}}

\usepackage{paralist}
\usepackage{verbatim}

\usepackage{newfloat}
\usepackage{listings}
\DeclareCaptionStyle{ruled}{labelfont=normalfont,labelsep=colon,strut=off} 
\floatstyle{ruled}
\newfloat{listing}{tb}{lst}{}
\floatname{listing}{Listing}
\graphicspath{ {./images/} }

\title{Attention-based Models for Snow-Water Equivalent Prediction}
\author {
    Krishu K Thapa\textsuperscript{\rm 1},
    Bhupinderjeet Singh \textsuperscript{\rm 1},
    Supriya Savalkar \textsuperscript{\rm 1},
    Alan Fern \textsuperscript{\rm 2},
    Kirti Rajagopalan\textsuperscript{\rm 1},
    Ananth Kalyanaraman\textsuperscript{\rm 1}
}
\affiliations {
    \textsuperscript{\rm 1}Washington State University, Pullman, WA.
    \textsuperscript{\rm 2}Oregon State University, Corvallis, OR.\\
    \{krishu.thapa, bhupinderjeet.singh, supriya.savalkar\}@wsu.edu,
    alan.fern@oregonstate.edu, \{kirtir, ananth\}@wsu.edu
}

\begin{document}

\maketitle

\begin{abstract}
Snow Water-Equivalent (SWE)---the amount of water available if snowpack is melted---is a key decision variable used by water management agencies to make irrigation, flood control, power generation and drought management decisions. SWE values vary spatiotemporally---affected by weather, topography and other environmental factors.
While daily SWE can be measured by Snow Telemetry (\snotel{}) stations with requisite instrumentation, such stations are spatially sparse requiring interpolation techniques to create spatiotemporally complete data. While recent efforts have explored machine learning (ML) for SWE prediction, a number of recent ML advances have yet to be considered. The main contribution of this paper is to explore one such ML advance, attention mechanisms, for SWE prediction. Our hypothesis is that attention has a unique ability to capture and exploit correlations that may exist across locations or the temporal spectrum (or both). We present a generic attention-based modeling framework for SWE prediction and adapt it to capture spatial attention and temporal attention. Our experimental results on 323 \snotel{} stations in the Western U.S. demonstrate that our attention-based models outperform other machine learning approaches. We also provide key results highlighting the differences between spatial and temporal attention in this context and a roadmap toward deployment for generating spatially-complete SWE maps.
\end{abstract}


\section{Introduction}
\label{sec:intro}


Much of the high-value, productive irrigated agriculture in the Western U.S. is facilitated by snowmelt-driven streamflow. Many of these watersheds are ``fully appropriated'' -- that is, every drop of water is appropriated to a particular use under a legal water rights system. Irrigation needs to be balanced with other uses such as hydropower production, municipal needs, and environmental flows. In managing these diverse uses in water-scarce environments, key decision variables used by  water management agencies include an expectation of the snowpack---measured as \emph{Snow Water-Equivalent (SWE)}---and the timing of snowmelt. SWE is the amount of water that would be available if the snowpack is melted. For example, the U.S. Bureau of Reclamation makes decisions on storage and release of water from reservoirs to meet diverse needs via forecasts of inflows into the reservoir, which uses SWE as a key input.

\begin{figure}[!tb]
\centering
\includegraphics[scale=0.30]{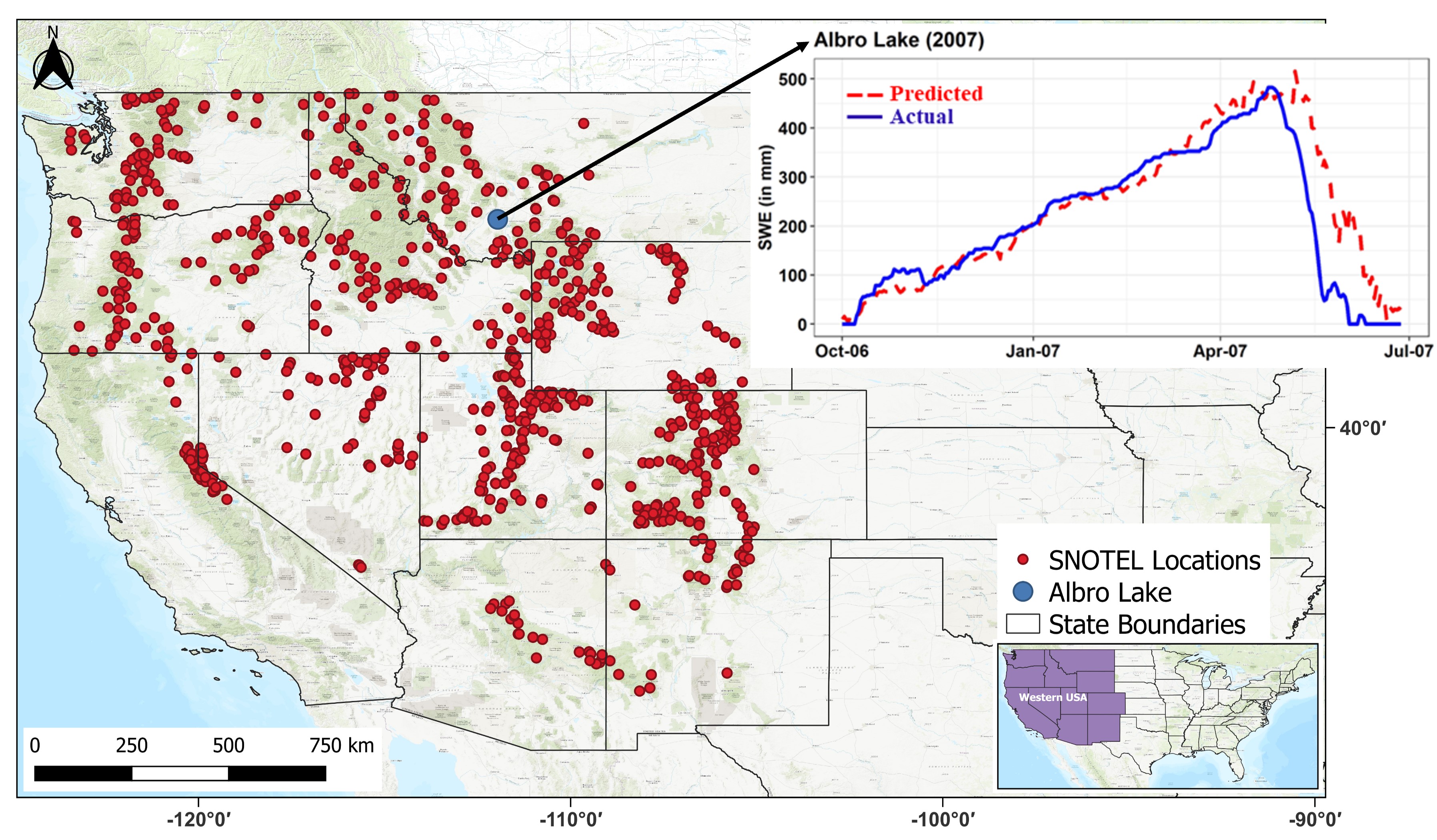}
\captionof{figure}{ \small Spatial map showing the sparse spread of  822 \snotel{} locations across Western U.S., of which we used a subset of 323 stations  based on data availability. The inset shows two SWE curves for a station (Albro Lake) for the 2007 SWE season. The blue curve is ground-truth (\snotel) and the red curve is the predicted curve obtained by our ensemble attention model. 
}
\label{fig:SWEcurve}
\end{figure}

    Given the importance of this information, over the last few decades, large investments were made in measurements via the Snow Telemetry (\snotel) station network across the Western U.S. These stations report the daily SWE value along with other related attributes including (but not limited to) snow depth, temperature, precipitation,  and relative humidity. However, these stations are sparsely distributed in space (as shown in the map of Figure~\ref{fig:SWEcurve}).
    The 822 \snotel{} stations shown, average to around one station for every 4,014 square kilometers.


    Currently, long-term spatially-complete SWE products are developed via interpolation from \snotel{} and other observational networks \cite{1c168b7f17c14c6b92e0df70cb4974a5}. 
    (For a brief review of current tools, see Section~\ref{sec:relatedworks}). These methods do not account for the large variations in SWE in short distances and take a simplistic approach to incorporating the effects of environmental variables.
    
    While \snotel{} information may be spatially-incomplete, there are other types of data  such as  remotely-sensed reflectances \cite{brodzik2016measures} or topographical details, and weather networks (providing temperature, precipitation etc.) that are denser than the \snotel{} network. Leveraging these data along with state-of-the-art AI models can significantly enhance our ability to obtain snowpack information at watershed scales and significantly improve decision-making. Such data can  advance hydrological models by reducing the uncertainties involved in their parameterization of snow processes.


 In this paper, we address the problem of predicting daily SWE for any location (with or without a \snotel{} station) on any given date, using observable data that are available at high spatial resolution (e.g., weather, satellite imagery).
Our prediction problem is motivated by the above established need for generating spatially-complete snowpack products.
Once such a generic prediction model is designed, it can also be extended to forecasting scenarios.

\vspace*{-0.05in}
\paragraph{Challenges:}
Predicting SWE for a particular location and time is a challenging task for multiple reasons. SWE is a complex trait affected by environmental variables and spatial attributes (e.g., elevation, latitude-longitude, topography, land cover)~\cite{sweEnsemble,1d35320e7f0947c7bd89ffd8c767d547,tempPrecipRef}.
Time is another dimension affecting prediction. The window of time from when the snow starts to accumulate to when snow melts completely is referred to as the \emph{SWE season}. While the SWE season varies geographically, an approximate 270-days window of (Oct. 1 to Jul. 1) covers the majority of the snow season for the Western U.S.
Figure~\ref{fig:SWEcurve} shows the general expected trend of the SWE curve over the SWE season for a \snotel{} station and year. 
However, there is significant heterogeneity in these curves across locations and seasons. For instance, the overall SWE values, maximum SWE and the date when the ground is snow-free, could all be different for a drought year versus a non-drought year.

Machine learning (ML) models  have been recently explored for SWE prediction \cite{bair2018using,Marofi2011PredictingSD,meyal2020automated}---see Section~\ref{sec:relatedworks}. These approaches pose the problem as time series prediction and use deep learning models such as recurrent neural networks.
However, they largely ignore spatial correlations.

\paragraph{Contributions:}
We explore \emph{attention} as the main mechanism  for  SWE prediction. Self-attention was introduced as part of the transformer architecture \cite{vaswani2017}. The main idea is to compute a set of attention weights for all token pairs  in a sequence and use those weights to make predictions. 
Even though originally applied to machine translation, it has since been widely adopted in other applications that involve a sequence-to-sequence transformation. 

\begin{figure}[tbh]
\centering
\includegraphics[scale=0.20]{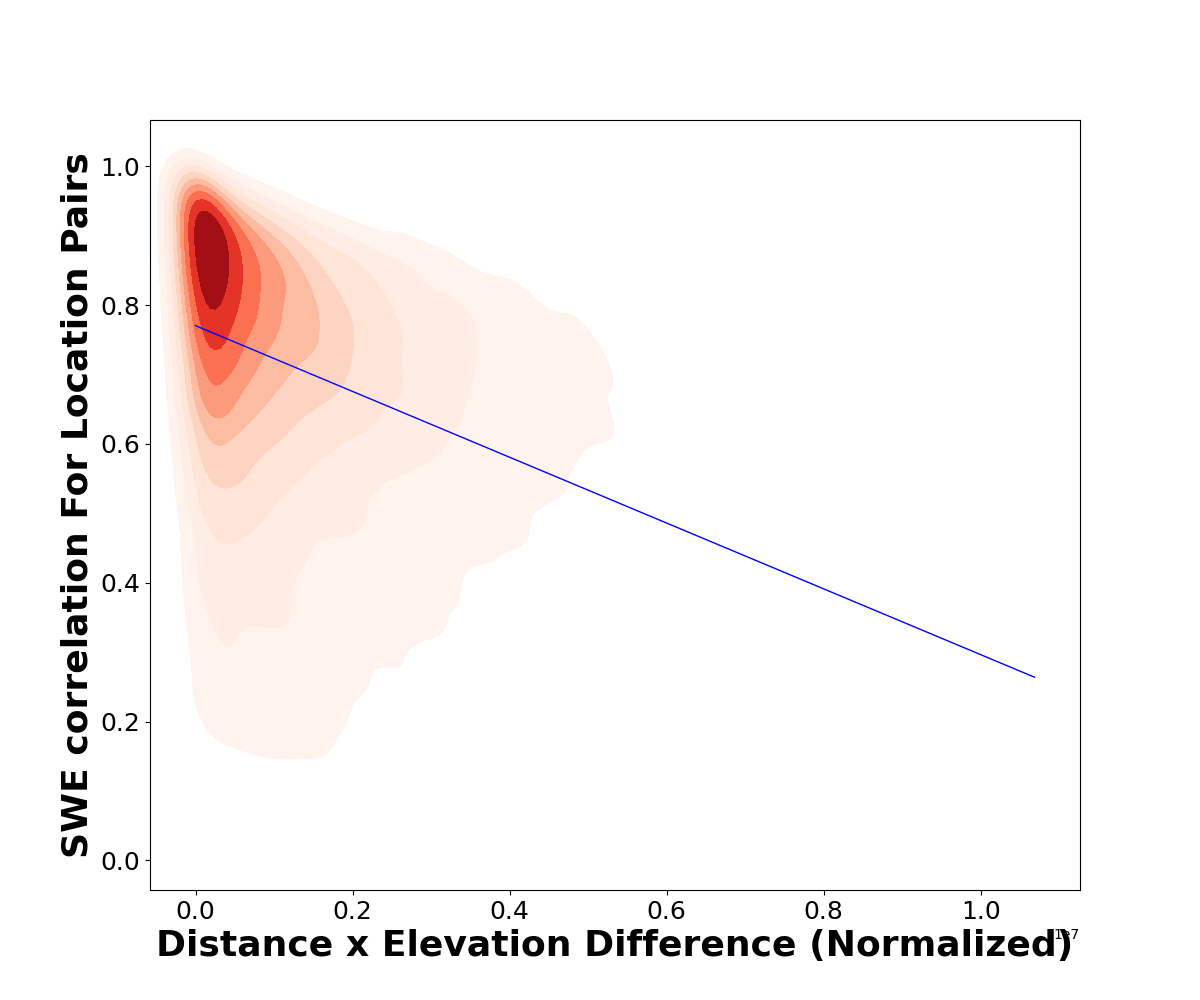}
\captionof{figure}{\small SWE correlation between all location pairs as a function of their spatial distance and elevation difference product. 
}
\label{fig:DistElevProduct}
\end{figure}

\emph{We hypothesize that the attention mechanism can help the SWE prediction task by capturing and exploiting any correlations that may exist between locations (spatial attention) or between time intervals (temporal attention).}
For instance, in a preliminary study using all \snotel{} stations in the  U.S., we observed strong correlations between  locations that have a short spatial distance or similar elevation or both (Figure~\ref{fig:DistElevProduct}). 

To  test our working hypothesis, we first designed a generic attention-based model architecture and adapted it to perform spatial attention, temporal attention, and a simple ensemble version which benefits from both.

Our experiments on 323 \snotel{} locations in the Western U.S. demonstrate the general effectiveness of the attention based schemes, which outperform other prior approaches. Our study also highlights the differences between the spatial and temporal attention schemes.  An extensive report on our experiments and findings are presented in the Section~\ref{sec:results}
, followed by a roadmap (Section~\ref{sec:roadmap}) toward deployment for generating a spatially-complete SWE map.

\section{Related work}
\label{sec:relatedworks}

Among the earlier work, \citet{HUANG1996159} developed a Kalman filter for predicting SWE. Their approach uses observed temperature and precipitation to simulate  snow cover conditions. 
The Kalman filter assumes that the relationship between the independent and the dependent variables is linear. However, this  assumption may or may not hold depending on the location and its various attributes including elevation and weather conditions. 
Other work used spatial interpolation techniques such as kriging \cite{carroll1999spatial} and
statistical regression tree \cite{elder1998estimating}. 
While these approaches use statistical models to interpolate by spatial proximity, they lack the sophistication to include additional features (e.g., weather, elevation, satellite reflectance observations) which could be key SWE indicators. 

Examples of spatially-complete SWE estimates include Barrett (2003) and Broxton et al. (2016). The former is based on integration of numerical weather models and data assimilation but is only available starting 2003. The latter is based on simple interpolation of \snotel{} and other observational data and is available as a longer time series. However spatiotemporal correlations are not accounted for, and the effect of other environmental variables are considered in a simplistic manner. Neither explored ML methods. 

ML approaches have been explored more recently.  \citet{bair2018using} used bagged trees and neural networks to predict SWE in the watersheds of Afghanistan. 
\citet{Marofi2011PredictingSD} use multivariate non-linear regression and artificial neural networks with kriging. 
\citet{meyal2020automated} present a recurrent neural network-based model using Long Short-Term Memory or LSTM \cite{hochreiter1997long} to predict SWE values in five \snotel{} locations in Colorado. 

In a concurrent work, \citet{duan2023using} (unpublished) compare multiple temporal deep learning models for SWE prediction, including LSTM, temporal convolutional neural network, and transformer architecture. 
In contrast to these efforts, \emph{this} paper presents a modeling framework to capture attention in both spatial and temporal dimensions. Our evaluations show that attention can be effective, outperforming other deep learning approaches.


\section{Approach}
\label{sec:approach}

Let $X$ denote the set  of all $n$ locations, and $m$ be the number of days in any SWE season (e.g., $\sim$270 days for Oct. 1 to July 1).
Let $S$ denote the set of all SWE seasons (i.e., years).
Let $x^{j,h}_i$ denote a vector of directly observable attributes for location $i\in X$ on day $j\in[1,m]$ of season $h\in S$.
Note that these attributes typically consist of static attributes (e.g., lat/long, elevation) and dynamic attributes (including daily weather variables, satellite reflectance observations). Finally, let $y^{j,h}_i$ be the corresponding daily SWE value, which is the primary prediction target in this work. 

The goal of the prediction problem is: given the collection of attribute vectors $\{x^{j,h}_i\}$, enumerated over locations, seasons, and days, predict the corresponding daily SWE value of each $y^{j,h}_i$. Note that an accurate solution to this problem allows for estimating any missing SWE value for a location, season, and day based on just observable attributes in the data without requiring any ground truth SWE. Training the model, however, does assume the availability of labeled SWE data.  

\paragraph{Spatial attention:}
We first consider our main ideas for implementing the spatial attention model. 
For mapping to the attention framework \cite{vaswani2017}, we reformulate the prediction problem as a sequence-to-sequence prediction:
\[ [y^{j,h}_{1}, y^{j,h}_{2},\ldots y^{j,h}_{n}]\gets f_{spatial}([x^{j,h}_{1},x^{j,h}_{2},\ldots x^{j,h}_{n}]) \]
where, $f_{spatial}$ is the function for the transformer architecture to learn as described in more detail in Section \ref{sec:model}. 

There are several advantages to this formulation:
\begin{itemize}\itemsep=-0.05ex
    \item First, the encoder within the transformer architecture can compute \emph{attention for every pair of locations} (for the same day), providing an opportunity to exploit any spatial correlation between location pairs.
    Intuitively, this also implies that if there are missing data for some locations then the model may still be able to learn from other locations that are sufficiently correlated to the missing locations.
    \item Second, although the attention mechanism works in the spatial dimension, one can also \emph{incorporate some temporal information}. 
More specifically, we divide each input location feature $x^{j,h}_i$ further into three parts i.e.;  $x^{j,h}_i$ = $[\phi , \alpha , \gamma]$. Here, $\phi$ represents the static spatial features like elevation, lat/long, etc.; $\alpha$ represents the daily observations for the location (including weather and satellite image features); and $\gamma$ represents the average of SWE observations on the same day $j$ but from a window of adjacent $w$ years, i.e $\gamma = \frac{1}{2w + 1} \sum_{t = h - w}^{h+w} x^{(j,t)}_{i} $.
Intuitively, the idea of $\gamma$ is to provide an added temporal context. 
\item There are other benefits to incorporating spatial attention. The set of \emph{attention vectors} computed in the final model can be of value in itself, as it could reveal interesting hidden relations between the locations. Furthermore, in this formulation, each training example is for a given $\langle$day $j$, season $h$$\rangle$ combination---which implies at test time we can use this model as an \emph{imputation tool} to fill SWE data for any location on arbitrary days of past SWE seasons.
\end{itemize}

\paragraph{Temporal attention:}
The formulation for temporal attention in SWE prediction can be achieved using a simple modification to the spatial attention formulation. 
Intuitively, instead of posing each input as a sequence of locations, we pose it as a sequence of ($m$) days for each location $i$: 
\[ [y^{1,h}_{i}, y^{2,h}_{i},\ldots y^{m,h}_{i}]\gets f_{temporal}([x^{1,h}_{i},x^{2,h}_{i},\ldots x^{m,h}_{i}]), \]
%
where $f_{temporal}$ is the transformer-structured function to learn.
Despite the similarity in the formulations, there are some subtle but important differences to the temporal attention scheme (compared to its spatial counterpart). 
\begin{itemize}\itemsep=-0.05ex
    \item First, unlike the spatial model (where we restrict $\alpha$ to a single day), here we have the opportunity to learn from a longer time window---e.g., a warmer-than-usual January and associated melt can change the snow density and age structure, which when followed by two cold months, can  impact the SWE in a following warmer month.
    In the absence of any such distinctive long-term correlations, however, this architecture can be expected to perform similar to an RNN  architecture such as LSTM.
    \item 
    Furthermore, since the temporal attention scheme is tied to a single location ($i$) during prediction,
    any pattern to missing data (particularly from consecutive time windows)  could potentially impact prediction accuracy. 
     
    
\end{itemize}


\paragraph{Ensemble attention:}
We can also combine the strengths of both attention mechanisms through ensemble methods. These methods can produce a net predicted value that could possibly surpass the prediction achieved individually. In principle, we can perform ensemble using any aggregator function $\oplus$:
{\[y^{j,h}_{i,ensemble} = y^{j,h}_{i,spatial}\oplus y^{j,h}_{i,temporal}\]}
%
We evaluate \textit{average} as a simple aggregator, i.e., the average of the two predictions is the ensemble prediction. More sophisticated ensemble schemes can be explored in the future.

\section{Model architecture}
\label{sec:model}
\begin{figure}[tbh]
\centering
        \centering
        \includegraphics[scale=0.5]{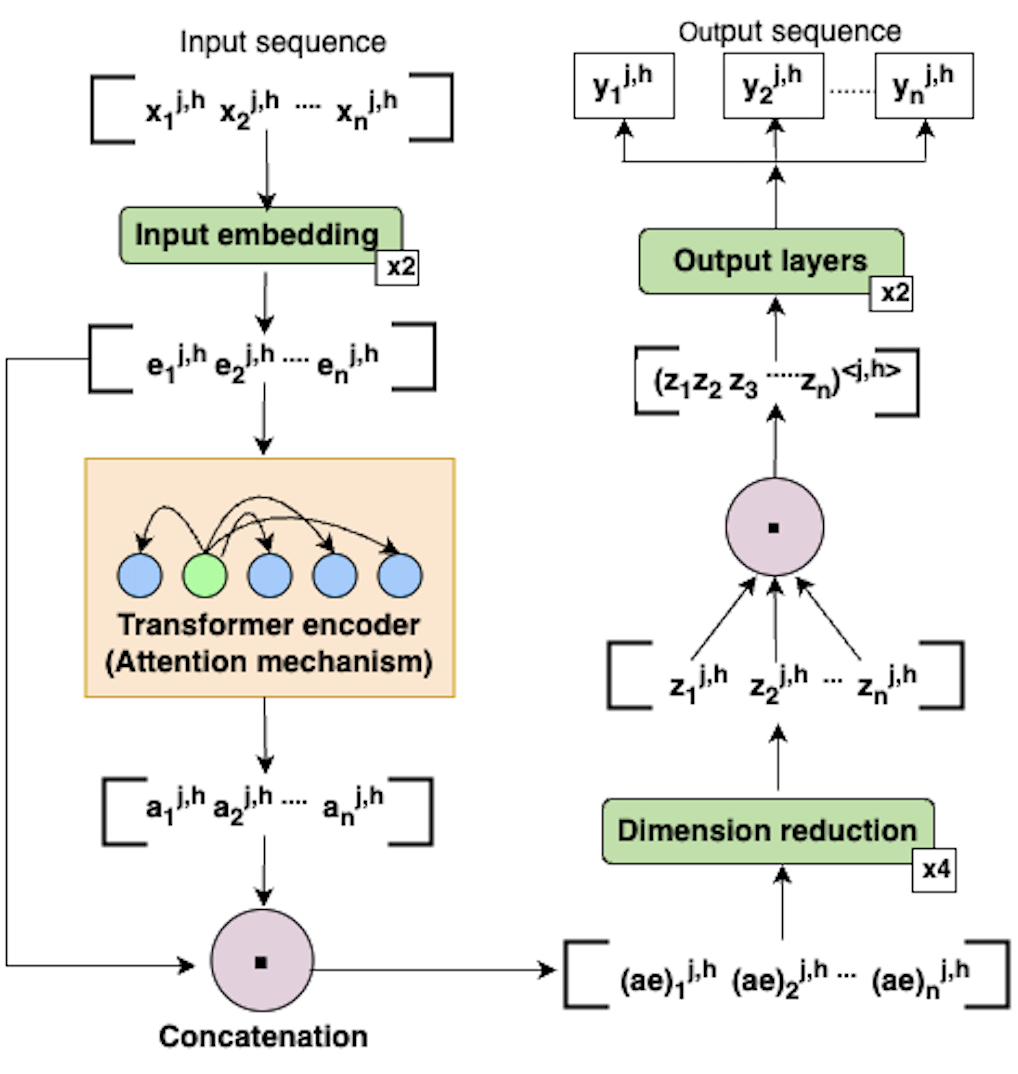}
        \caption{\small
        Spatial attention model architecture. This is easily adapted to also implement our temporal attention model. 
        }
        \label{fig: modelArchitecture}
        
 \end{figure}

We adapt the transformer model architecture \cite{vaswani2017} to implement spatial and temporal attention. 
For ease of exposition, we describe the model architecture for the spatial  scheme. Temporal attention can be implemented in the same architecture using a trivial modification to the inputs and outputs (as described above). Figure~\ref{fig: modelArchitecture} 
 shows our transformer architecture for the spatial attention model.
The model uses the classical transformer encoder  \cite{vaswani2017}  to encode attention between every pair of locations. 
The encoded information is then passed to a series of linear layers to obtain the SWE value for each location.


\vspace*{-0.05in}
\paragraph{Input:}
Each training example is a sequence of locations $[x^{j,h}_{1},x^{j,h}_{2},\ldots x^{j,h}_{n}]$  for day $j$ and season $h$.
We denote this sequence as \inputseqx.

\vspace*{-0.05in}
\paragraph{Output:}
The output of the model is the sequence of predicted daily SWE values  $[y^{j,h}_{1}, y^{j,h}_{2},\ldots y^{j,h}_{n}]$ for all locations for day $j$ and season $h$.
We denote this sequence as \outputseqy.


\begin{compactenum}
    \item{\bf{Input embedding layers}:}
    Two embedding linear layers are used to generate a $d$-dimensional embedding (\embedseqe) for the input sequence \inputseqx. We used $d=512$ in our experiments. 
    Both layers use Gaussian Error Linear Unit (GELU) for activation. 
    \item{\bf{Transformer encoder}:}
    For self-attention, we used 16 attention heads and 24 feed-forward layers within the block, to generate the encoded sequence (\encodseqa), which is 
    $a_1^{j,h}, a_2^{j,h},\ldots, a_n^{j,h}$. 
    
    \item{\bf{Locationwise concatenation}:}
    Next, the encoded vector \encodseqa{} is concatenated with the embedded vector \embedseqe, using a location-wise concatenation operation, represented by \concatseqae: $(ae)_1^{j,h}, (ae)_2^{j,h},\ldots,(ae)_n^{j,h}$. The rationale behind this concatenation is to carry forward the initial input representation along with its encoded representation. 
    
    \item{\bf{Dimension reduction layers}:}
    Note that the above step would have doubled the number of dimensions due to the concatenation (i.e., $d=1024$).
    Subsequently, each location's representation in the sequence is passed through four linear layers to reduce its dimensionality by a factor of 8---i.e., if $d=1024$ at the start of this step, it becomes $128$ at the end. 
    Furthermore, within each layer, a dropout with a rate of 0.20 is applied. The first and third layers were subjected to the Rectified Linear Unit (RELU) as an activation function. The dimension-reduced sequence is denoted as \reduceddimz: $z_1^{j,h}, z_2^{j,h},\ldots, z_n^{j,h}$.
    
    \item {\bf All locations concatenation}:
    In this block, all elements in the sequence \reduceddimz{} are concatenated to form a single large vector of features, represented by $(z_1z_2z_3....z_n)^{j,h}$. The rationale for this concatenation is that during the output extraction for an individual location, the output block will have an opportunity to make use of both its own location's representation as well as others. 
    
    \item{\bf{Output layers}:}
    This block contains two linear layers that transform the concatenated representation to the final prediction for each of the $n$ locations. 
    The first layer uses GELU activation and has a dropout rate of 0.10. 
    The second layer uses the identity activation function to obtain the daily SWE values (\outputseqy).
    \item{\bf Loss function:}
     We used Mean Squared Error as the loss function, the AdamW Optimizer, and an initial learning rate of 0.0001. A scheduler reduces the learning rate by factor of 0.6 after every three epochs during training.
\end{compactenum}

\vspace*{-0.05in}
\paragraph{Implementation and availability:}
Model implementation used the Python Scikit-learn {\textit{(v1.1.2)}} ( Linear Regression), and Pytorch {\textit{(v2.0.1)}} (LSTM and Attention models) packages. Data processing and visualization used multiple Python and R packages. 
Code and data are accessible at: \url{https://github.com/Krishuthapa/SWE-Attention}.

\section{Experimental setup}
\label{sec:ExpSetup}

\paragraph{Data description:}
\label{sec:Data}
The set of static and dynamic features used along with their respective sources are listed below:
\begin{itemize}\itemsep=-0.05ex
   \item \textbf{Static features}:
     elevation,latitude, longitude  \cite{NRCS:2023}; land cover \cite{Yang_2018}; southness 
     \cite{NED:2014}.  
    \item{\bf Dynamic (daily) \snotel{} features:}
        SWE, precipitation, minimum, maximum, and average temperatures \cite{NRCS:2023}
    \item{\bf Dynamic (daily) satellite observations:}
        Passive microwave brightness temperature (19VE, 37VE, and their difference)\cite{brodzik2016measures}
 \end{itemize}

The daily \snotel{} data consists of 822 stations for 18 water years (2002-2019).
We filtered out stations with more than 10\% missing values in any variable or year. The resulting 323 stations comprised our main data set. Given our focus on SWE predictions, we used daily data for a 270 day-period starting Oct 1 as representative of the SWE season. This resulted in a total of 1,569,780 (=$323\times 18\times 270$) $\langle$location, year, day$\rangle$ combinations.  

The satellite observations were available at spatial resolutions of 6.25 km (19 GHz) and 3.125 km (37 GHz). The 3.125 km resolution data was resampled to 6.25 km using a nearest neighbor interpolation. Values for a location were assigned from the grid containing it. The landcover grid was at a 30m resolution and the dominant landcover type from the grid containing each location was extracted. A Digital Elevation Model 30m grid derived from the LIDAR point cloud source was processed in ArcGIS to obtain slope and elevation. From this southness was calculated as 
     $\cos(\textrm{aspect})\cdot \sin(\textrm{slope})$

\paragraph{Evaluation methodology:}
\label{sec:evaluationmethodology}

We split the 18 years into two sets: training (13), and testing (5). 
The five testing years  cover the entire spectrum of average annual SWE. Test water years are 2015, 2007, 2018, 2017, and 2008---ordered from the lowest to highest average annual SWE.


 In our experiments, we compared five models:
 \begin{itemize}\itemsep=-0.05ex
     \item{\spaceatt:} the proposed spatial attention model
     \item{\temporalatt:} the proposed temporal attention model
     \item{\ensembleatt:}  the proposed ensemble attention model which uses a simple average between the spatial and temporal attention predictions
     \item{LSTM:} Long Short-Term Memory model  \cite{hochreiter1997long}
     \item{LR:} Linear regression
 \end{itemize}

 To compare models, we use the \emph{Nash-Suttcliffe Efficiency (NSE)} metric \cite{nash1970river}, a widely adopted standard for measuring the accuracy of hydrological models. NSE for each location $i\in X$ is calculated as follows:
 \begin{equation}\label{eqn:NSEformula}
 NSE_{i}  =  1 - \frac{\sum_{t=1}^{T} (Y_{a,i}^{t} - Y_{p,i}^{t})^{2}}{\sum_{t=1}^{T} (Y_{a,i}^{t} - \bar{Y}_{a,i})^{2}}     
 \end{equation}
where, $T$ is the total number of distinct year-month-days used in testing for SWE prediction,  $Y^t_{a,i}$ and $Y^t_{p,i}$ are actual and predicted SWE respectively for a given day $t\in[1,T]$, and  $\bar{Y}_{a,i}$ is the mean of all the actual SWE values for location $i$ over all $T$ days.
 The value of NSE ranges from $-\infty$ to 1, with a value closer to 1 implying the best performance, and $\leq 0$ implying that prediction is worse than using the long-term mean as the prediction.
 In practice, a value of 0.5 or above is subjectively considered good. We also calculated mean errors between the predicted and observed daily SWE values and annual maximum SWE values.

\section{Experimental results}
\label{sec:results}

\begin{figure}[tb]
\centering
\includegraphics[scale=0.33]{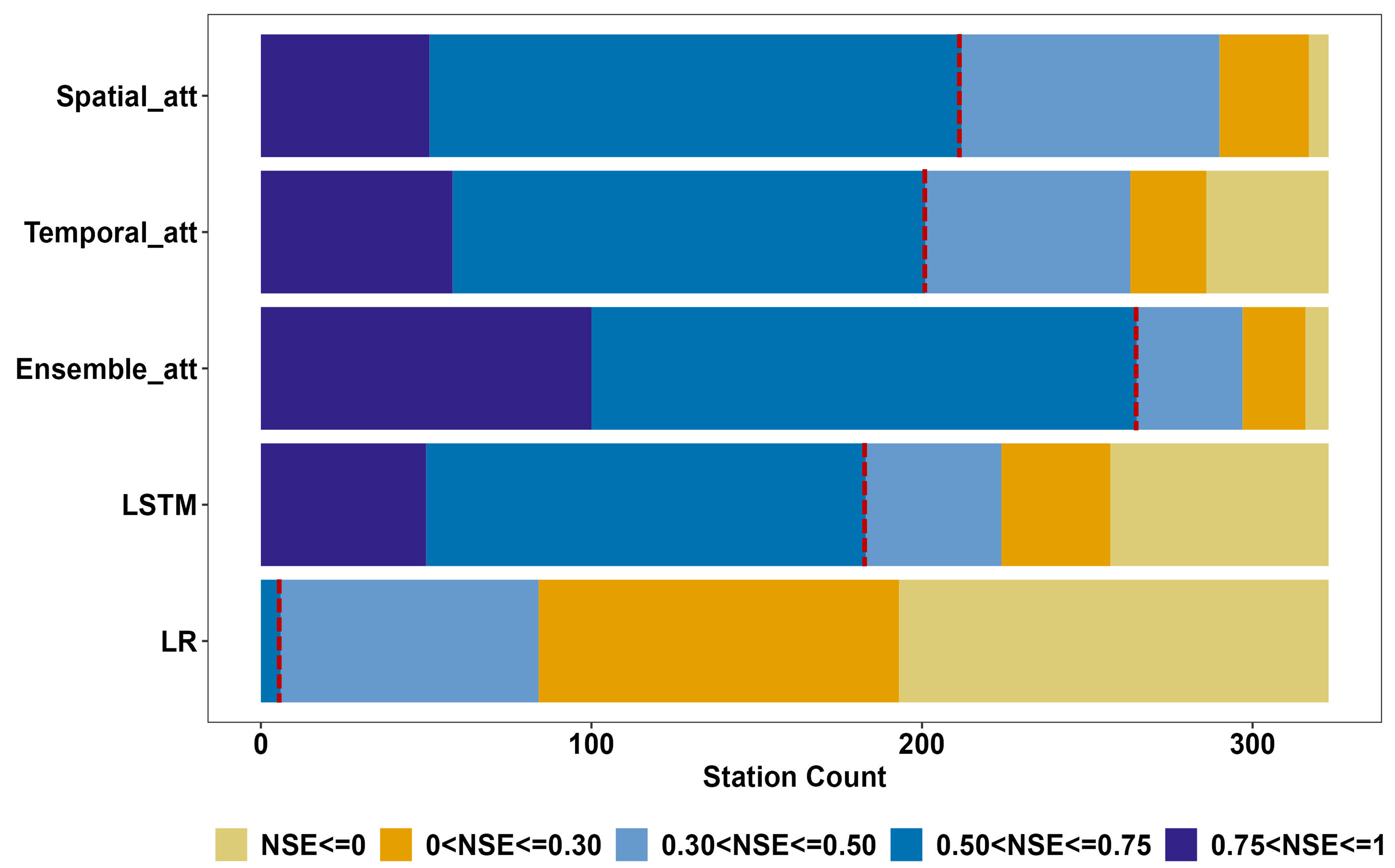}
\caption{\small The distribution of locations across five NSE groups for all models. The NSE metric is calculated for each location, with higher values (blue bars) indicating a better prediction. The left of the red dotted line corresponds to NSE $>$ 0.5.}
\label{fig:nseALL}
\end{figure}

\begin{figure}[h!]
\centering
\includegraphics[scale=0.45]{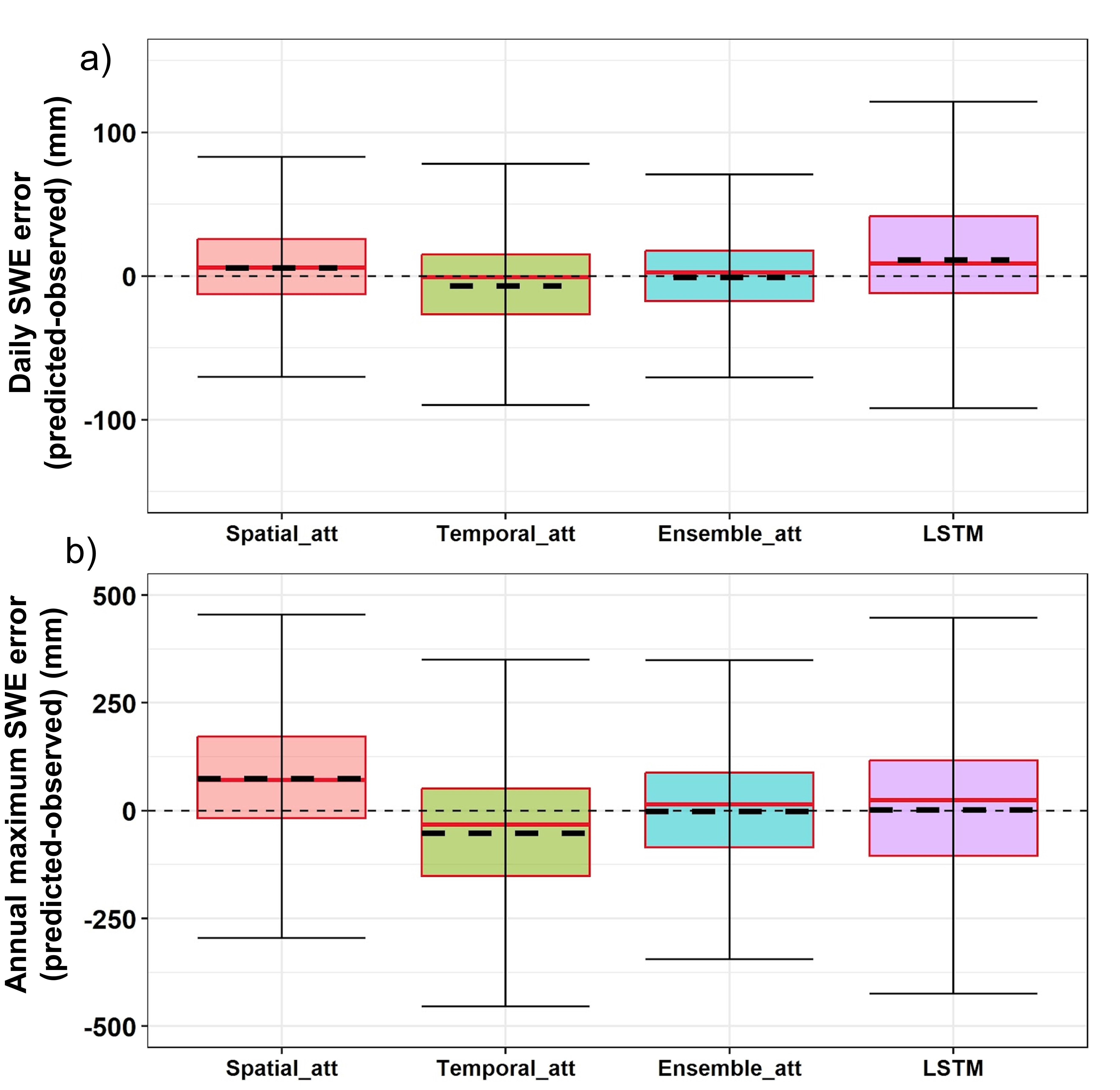}
\caption{\small Mean error \emph{(predicted - observed)} in SWE (mm) for each location. Part (a) shows the errors in daily SWE across the 270 days of a SWE season. The range in each box plot corresponds to errors from 270 days $\times$ 323 locations. Part (b) shows a similar plot but for the annual maximum SWE predictions.}
\label{fig:maxError}
\end{figure}

\paragraph{Model comparison:}

We first analyze the performance of models by their NSE values for the test dataset.  NSE values were calculated for each location using Eq.~\ref{eqn:NSEformula}.
Figure~\ref{fig:nseALL} shows the distribution of all locations binned into different NSE groups. Observations are as follows.

\begin{itemize}\itemsep=-0.05ex
    \item All three attention models outperform the other two  models (LSTM and LR). 
    \item Among the attention models, \ensembleatt{} is the best, followed closely by \spaceatt{} and then \temporalatt. For example, \ensembleatt{} predictions placed {82.04\%} of the locations with NSE values $>0.50$; with the corresponding fractions being {65.63\%} for \spaceatt{} and {62.22\%} for \temporalatt{}.
   
    \item \temporalatt{} performs slightly better than LSTM. Even though both use temporal information, LSTM processes  the sequence sequentially, while the attention mechanism has the freedom to derive context from either end of the sequence.
\end{itemize}

\begin{figure*}[tbh]
    \centering
    \includegraphics[scale=0.47]{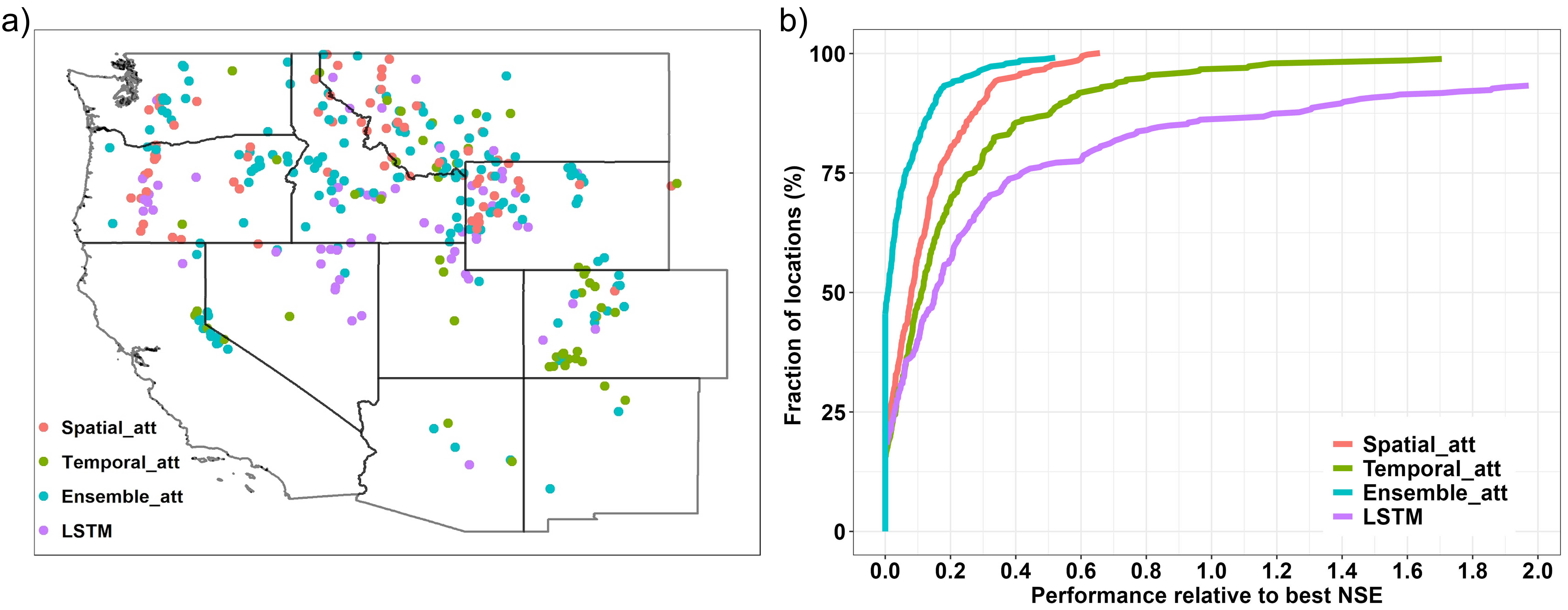}
    \caption{\small
    a) The best model (highest NSE) for each location overlaid on Western U.S. map. 
    b) Relative model performance (RMP) chart: A model's RMP is measured as the difference in its NSE from the best NSE for each location (X-axis). For visual clarity, the X-axis is restricted to a RMP of 2.0. This covers 94\% of the data and primarily excludes LSTM data, resulting in the purple line not reaching a 100\% fraction.The Y-axis is the fraction of locations for which that RMP is achieved. The closer a curve is to the Y-axis and for longer, the better.  
    }
    \label{fig:spatialPerformance}
\end{figure*}

Better performance of the \ensembleatt{} is partially due to the respective under- and over- prediction by the \spaceatt{} and \temporalatt{} models (Figure~\ref{fig:maxError}a), resulting in an ensemble with a higher accuracy and lower spread in errors. This behavior is pronounced when comparing the annual max SWE (Figure~\ref{fig:maxError}b). 

While the \ensembleatt{} model works best in most locations, we observed a spatiotemporal variation in the best-performing model (Figure~\ref{fig:spatialPerformance}a). 
To examine model performance at a finer level, we plotted a relative performance chart shown in Figure~\ref{fig:spatialPerformance}b. For each location, we recorded the difference of the NSE obtained by a given model from the NSE for the best performing model for that location. The X-axis shows this difference as the ``relative performance'' (0 is best). We then plotted the fraction of locations (Y-axis) for which each model produced a given relative performance.  
In this representation, the closer a model's curve aligns with the Y-axis the better.
The best model, \ensembleatt,  achieves the best NSE value for around 50\% of the locations. Notably, even in cases where it is not the best, its difference from the best is at most 0.2 for almost 90\% of the location.
\spaceatt{} also provided a comparable performance. 

Some of the spatial variation is attributable to differential performance for stations at different \emph{elevations}---an important factor affecting snow dynamics.
Table~\ref{tab:elevTable} shows the median NSE values within each elevation group for each model. 
In the lower elevation areas, which  are generally harder to predict, we found that \spaceatt{} performed significantly better. 
\temporalatt{} performed the best at the highest elevation group, and the \ensembleatt{} performed best in the mid-range elevation groups.

\begin{table}[htb]
\captionsetup{singlelinecheck=false}
\centering
\begin{tabular}{ccccc}
\bottomrule
\rowcolor{shadecolor}  
      { \small Elevation }&
      { \small Spatial}&
      { \small Temporal}&
      { \small Ensemble}&
      { \small LSTM} \\
\rowcolor{shadecolor}  
      { \small  Group}&
      { \small Att.}&
      { \small Att.}&
      { \small Att.}&
      { \small } \\
      
\hline
\centering
{\small [0\%, 25\%)} & {\small \bf{0.50}} & {\small 0.24} &{\small 0.49} & {\small 0.31} \\
{\small [25\%, 50\%)} & {\small 0.48} & {\small 0.64} & { \small \textbf{0.70}} &{ \small 0.48} \\
{\small [50\%, 75\%)} & {\small 0.61} & {\small 0.66} & { \small \textbf{0.72}} &{ \small 0.59} \\
{\small [75\%, 100\%)} & {\small 0.52} & {\small \textbf{0.69}} & {\small 0.68} &{ \small 0.59} \\
\hline
\toprule
\end{tabular}
\caption{\small Median NSE of each model across locations for four elevation groups. The groups are based on quartiles of elevations, and the rows are sorted from low to high elevation.
Boldface entries have the highest median NSE for each group (row).}
\label{tab:elevTable}
\end{table}

\vspace*{-0.05in}
\paragraph{Model run-times:}
 The models were trained on a 3.1GHz Apple M1 Pro processor with 16 GB RAM. 
 The training times for the different models were: \spaceatt{} (440 min); \temporalatt{} (450 min); and LSTM (464 min). 
Note that the simple average ensemble can directly use the prediction outputs from both attention models.

\section{Roadmap to deployment}
\label{sec:roadmap}
The objective of this work was to lay the groundwork for a generic and extensible attention-based ML modeling framework for SWE prediction. One of the main drivers behind this work is to develop capability to generate spatially-complete SWE maps at country-scales. Such maps can become an integral part of snow products useful to hydrologists, irrigation district managers, and other resource planning commissions and agencies such as the United States Bureau of Reclamation (USBR). 

To enable this deployment, several extensions and evaluations will be needed including (but not limited to):
a) application as a spatial imputation tool to fill SWE values for non-SNOTEL locations across past decades (so as to generate an extensive gridded product); and
b) enable a forecast function (with uncertainty quantification) for prediction of future snowpack.
Ideally such deployment should happen in close partnership with the public sector (e.g., a government agency or service center such as USDA NRCS or USBR). 

Furthermore, this research has opened up multiple avenues of interesting research directions such as:
i) non-trivial ensemble methods,
ii) alternate ways to model the problem including spatiotemporal graph neural networks,
or alternatively, use the attention scheme to infer a graph representation,
iii) characterize the role of different variables, and
iv) couple with process-based models to ensure scientific consistency and model interpretability.




\section*{Acknowledgments}
\label{sec:ack}

This research was supported by USDA NIFA award No.
2021-67021-35344 (AgAID AI Institute). 
We thank Hossein Noorazar for help with some of the preliminary studies. 


\bibliography{references.bib}

\begin{thebibliography}{18}
\providecommand{\natexlab}[1]{#1}

\bibitem[{Bair et~al.(2018)Bair, Abreu~Calfa, Rittger, and Dozier}]{bair2018using}
Bair, E.~H.; Abreu~Calfa, A.; Rittger, K.; and Dozier, J. 2018.
\newblock Using machine learning for real-time estimates of snow water equivalent in the watersheds of Afghanistan.
\newblock \emph{The Cryosphere}, 12(5): 1579--1594.

\bibitem[{Brodzik et~al.(2016)Brodzik, Long, Hardman, Paget, and Armstrong}]{brodzik2016measures}
Brodzik, M.; Long, D.; Hardman, M.; Paget, A.; and Armstrong, R. 2016.
\newblock MEaSUREs calibrated enhanced-resolution passive microwave daily EASE-grid 2.0 brightness temperature ESDR, version 1.
\newblock \emph{Digital Media}.

\bibitem[{Broxton, Dawson, and Zeng(2016)}]{1c168b7f17c14c6b92e0df70cb4974a5}
Broxton, P.; Dawson, N.; and Zeng, X. 2016.
\newblock Linking snowfall and snow accumulation to generate spatial maps of SWE and snow depth.
\newblock \emph{Earth and Space Science}, 3(6): 246--256.

\bibitem[{Carroll, Carroll, and Poston(1999)}]{carroll1999spatial}
Carroll, S.~S.; Carroll, T.~R.; and Poston, R.~W. 1999.
\newblock Spatial modeling and prediction of snow-water equivalent using ground-based, airborne, and satellite snow data.
\newblock \emph{Journal of Geophysical Research: Atmospheres}, 104(D16): 19623--19629.

\bibitem[{Duan et~al.(2023)Duan, Ullrich, Risser, and Rhoades}]{duan2023using}
Duan, S.; Ullrich, P.; Risser, M.; and Rhoades, A. 2023.
\newblock Using Temporal Deep Learning Models to Estimate Daily Snow Water Equivalent over the Rocky Mountains.
\newblock \emph{Authorea Preprints (under review)}.

\bibitem[{Elder, Rosenthal, and Davis(1998)}]{elder1998estimating}
Elder, K.; Rosenthal, W.; and Davis, R.~E. 1998.
\newblock Estimating the spatial distribution of snow water equivalence in a montane watershed.
\newblock \emph{Hydrological Processes}, 12(10-11): 1793--1808.

\bibitem[{Hochreiter and Schmidhuber(1997)}]{hochreiter1997long}
Hochreiter, S.; and Schmidhuber, J. 1997.
\newblock Long short-term memory.
\newblock \emph{Neural computation}, 9(8): 1735--1780.

\bibitem[{Huang and Cressie(1996)}]{HUANG1996159}
Huang, H.-C.; and Cressie, N. 1996.
\newblock Spatio-temporal prediction of snow water equivalent using the Kalman filter.
\newblock \emph{Computational Statistics \& Data Analysis}, 22(2): 159--175.

\bibitem[{Kim et~al.(2021)Kim, Kumar, Vuyovich, Houser, Lundquist, Mudryk, Durand, Barros, Kim, Forman, Gutmann, Wrzesien, Garnaud, Sandells, Marshall, Cristea, Pflug, Johnston, Cao, and Wang}]{sweEnsemble}
Kim, R.~S.; Kumar, S.; Vuyovich, C.; Houser, P.; Lundquist, J.; Mudryk, L.; Durand, M.; Barros, A.; Kim, E.; Forman, B.; Gutmann, E.; Wrzesien, M.; Garnaud, C.; Sandells, M.; Marshall, H.-P.; Cristea, N.; Pflug, J.; Johnston, J.; Cao, Y.; and Wang, S. 2021.
\newblock Snow Ensemble Uncertainty Project (SEUP): Quantification of snow water equivalent uncertainty across North America via ensemble land surface modeling.
\newblock \emph{The Cryosphere}, 15: 771--791.

\bibitem[{Marofi, Tabari, and Abyaneh(2011)}]{Marofi2011PredictingSD}
Marofi, S.; Tabari, H.; and Abyaneh, H.~Z. 2011.
\newblock Predicting Spatial Distribution of Snow Water Equivalent Using Multivariate Non-linear Regression and Computational Intelligence Methods.
\newblock \emph{Water Resources Management}, 25: 1417--1435.

\bibitem[{Meyal et~al.(2020)Meyal, Versteeg, Alper, Johnson, Rodzianko, Franklin, and Wainwright}]{meyal2020automated}
Meyal, A.~Y.; Versteeg, R.; Alper, E.; Johnson, D.; Rodzianko, A.; Franklin, M.; and Wainwright, H. 2020.
\newblock Automated cloud based long short-term memory neural network based SWE prediction.
\newblock \emph{Frontiers in Water}, 2: 574917.

\bibitem[{Nash and Sutcliffe(1970)}]{nash1970river}
Nash, J.~E.; and Sutcliffe, J.~V. 1970.
\newblock River flow forecasting through conceptual models part I—A discussion of principles.
\newblock \emph{Journal of Hydrology}, 10(3): 282--290.

\bibitem[{NED(2014)}]{NED:2014}
NED, U. 2014.
\newblock United States Geological Survey, National Elevation Dataset (https://apps.nationalmap.gov/datasets/).
\newblock Accessed: November 3, 2023.

\bibitem[{NRCS(2023)}]{NRCS:2023}
NRCS. 2023.
\newblock USDA NRCS NWCC Database for SNOTEL locations data (https://wcc.sc.egov.usda.gov/reportGenerator/).
\newblock Accessed: November 3, 2023.

\bibitem[{Rutter et~al.(2009)Rutter, Essery, Pomeroy, and Altimir}]{1d35320e7f0947c7bd89ffd8c767d547}
Rutter, N.; Essery, R.; Pomeroy, J.; and Altimir, N. 2009.
\newblock Evaluation of forest snow processes models (SnowMIP2).
\newblock \emph{Journal of Geophysical Research: Atmospheres}, 114(D6).

\bibitem[{Sospedra-Alfonso, Melton, and Merryfield(2015)}]{tempPrecipRef}
Sospedra-Alfonso, R.; Melton, J.; and Merryfield, W. 2015.
\newblock Effects of temperature and precipitation on snowpack variability in the Central Rocky Mountains as a function of elevation.
\newblock \emph{Geophysical Research Letters}, 42: 4429--4438.

\bibitem[{Vaswani et~al.(2017)Vaswani, Shazeer, Parmar, Uszkoreit, Jones, Gomez, Kaiser, and Polosukhin}]{vaswani2017}
Vaswani, A.; Shazeer, N.; Parmar, N.; Uszkoreit, J.; Jones, L.; Gomez, A.~N.; Kaiser, L.~u.; and Polosukhin, I. 2017.
\newblock Attention is All you Need.
\newblock In Guyon, I.; Luxburg, U.~V.; Bengio, S.; Wallach, H.; Fergus, R.; Vishwanathan, S.; and Garnett, R., eds., \emph{Advances in Neural Information Processing Systems}, volume~30. Curran Associates, Inc.

\bibitem[{Yang et~al.(2018)Yang, Jin, Danielson, Homer, Gass, Bender, Case, Costello, Dewitz, Fry, Funk, Granneman, Liknes, Rigge, and Xian}]{Yang_2018}
Yang, L.; Jin, S.; Danielson, P.; Homer, C.; Gass, L.; Bender, S.~M.; Case, A.; Costello, C.; Dewitz, J.; Fry, J.; Funk, M.; Granneman, B.; Liknes, G.~C.; Rigge, M.; and Xian, G. 2018.
\newblock A new generation of the United States National Land Cover Database: Requirements, research priorities, design, and implementation strategies.
\newblock \emph{{ISPRS} Journal of Photogrammetry and Remote Sensing}, 146: 108--123.

\end{thebibliography}

\end{document}